\documentclass[runningheads]{llncs}
\usepackage{graphicx}
\usepackage{color}
\usepackage{amsmath}
\usepackage{amsfonts}

\begin{document}
\title{A Literature Review of Recent Graph Embedding Techniques for Biomedical Data}

\author{Yankai Chen\inst{1} \and
Yaozu Wu\inst{2} \and
Shicheng Ma\inst{2} \and
Irwin King \inst{1}}

\institute{Department of Computer Science and Engineering, \\ 
The Chinese University of Hong Kong \\
\email{\{ykchen,king\}@cse.cuhk.edu.hk} \\
\and KEEP, The Chinese University of Hong Kong\\
\email{yaozuwu279@gmail.com, shicheng@keep.edu.hk}}

\maketitle              
\begin{abstract}
With the rapid development of biomedical software and hardware, a large amount of relational data interlinking genes, proteins, chemical components, drugs, diseases, and symptoms has been collected for modern biomedical research. Many graph-based learning methods have been proposed to analyze such type of data, giving a deeper insight into the topology and knowledge behind the biomedical data, which greatly benefit to both academic research and industrial application for human healthcare. However, the main difficulty is how to handle high dimensionality and sparsity of the biomedical graphs. Recently, graph embedding methods provide an effective and efficient way to address the above issues. It converts graph-based data into a low dimensional vector space where the graph structural properties and knowledge information are well preserved. In this survey, we conduct a literature review of recent developments and trends in applying graph embedding methods for biomedical data. We also introduce important applications and tasks in the biomedical domain as well as associated public biomedical datasets. 
\end{abstract}
\section{Introduction}

With the recent advances in biomedical technology, a large number of relational data interlinking biomedical components including proteins, drugs, diseases, and symptoms, etc. has gained much attention in biomedical academic research. Relational data, also known as the graph, which captures the interactions (i.e., edges) between entities (i.e., nodes), now plays a key role in the modern machine learning domain. Analyzing these graphs provides users a deeper understanding of topology information and knowledge behind these graphs, and thus greatly benefits many biomedical applications such as biological graph analysis~\cite{albert2005scale}, network medicine~\cite{barabasi2011network}, clinical phenotyping and diagnosis~\cite{shickel2017deep}, etc.

\begin{figure}[tp]
\centering
     \includegraphics[width=1\columnwidth]{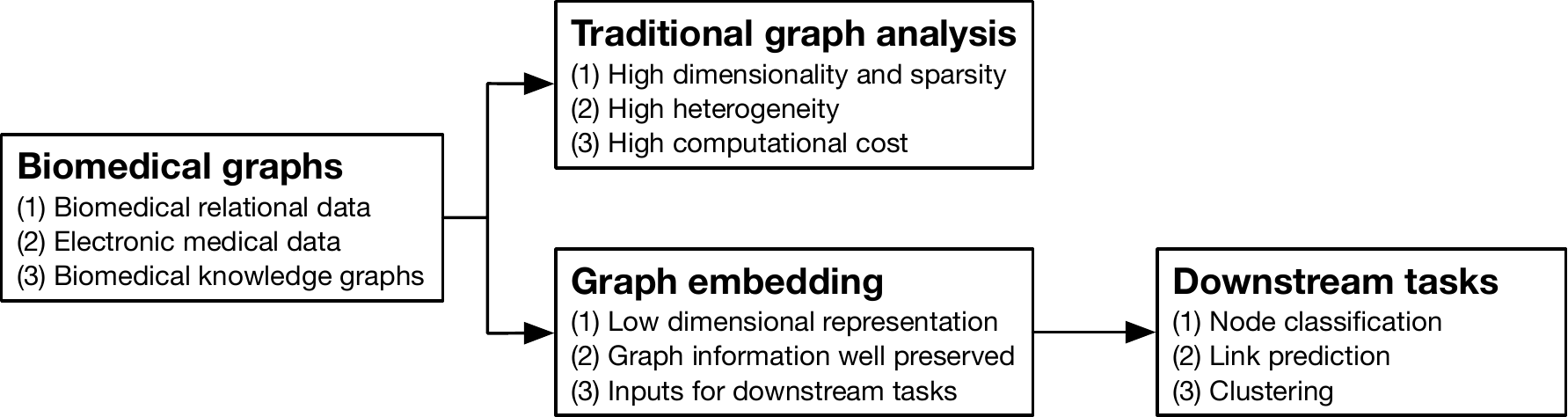}
    \caption{Comparison between traditional graph analysis methods and graph embedding techniques for biomedical graphs.}
    \label{fig:intro}
\end{figure}

As summarized in Figure~\ref{fig:intro}, although graph analytics is of great importance, most existing graph analytics methods suffer the computational cost drawn by high dimensionality and sparsity of the graphs~\cite{chen2020efficient,cai2018comprehensive,perozzi2014deepwalk}. Furthermore, owing to the heterogeneity of biomedical graphs, i.e., containing multiple types of nodes and edges, traditional analyses over biomedical graphs remain challenging.
Recently, graph embedding methods, aiming at learning a mapping that embeds nodes into a low dimensional vector space $\mathbb{R}^d$, now provide an effective and efficient way to address the problems. Specifically, the goal is to optimize this mapping so that the node representation in the embedding space can well preserve information and properties of the original graphs. After optimization of such representation learning, the learned embedding can then be used as feature inputs for many machine learning downstream tasks, which hence introduces enormous opportunities for biomedical data science. Efforts of applying graph embedding over biomedical data are recently made but still not thoroughly explored; capabilities of graph embedding for biomedical data are also not extensively evaluated. In addition, the biomedical graphs are usually sparse, incomplete, and heterogeneous, making graph embedding more complicated than other application domains. To address these issues, it is strongly motivated to understand and compare the state-of-the-art graph embedding techniques, and further study how these techniques can be adapted and applied to biomedical data science. Thus in this survey, we investigate recent developments and trends of graph embedding techniques for biomedical data, which give us better insights into future directions. In this article, we introduce the general models related to biomedical data and omit the complete technical details. For a more comprehensive overview of graph embedding techniques and applications, we refer readers to previous well-summarized papers~\cite{cai2018comprehensive,goyal2018graph,su2020network,chami2020machine}.

In this article, we first give the preliminaries used in this paper. We then briefly introduce the widely used graph embedding models. After that, we introduce some related public biomedical datasets. Finally, we carefully discuss the recent developments and trends of biomedical graph embedding applications.

\section{Preliminaries}
\vspace{-0.1in}





\begin{definition}[Homogeneous graphs]
A homogeneous graph $G = (V, E)$ is associated with two mapping functions $\Phi: V$ (node set) $\rightarrow \mathcal{A}$ (node type set) and $\Psi: E$ (edge set) $\rightarrow \mathcal{R}$ (edge type set) and $|\mathcal{A}| = |\mathcal{R}| = 1$. 
\end{definition}
\vspace{-0.1in}

\begin{definition}[Heterogeneous graphs]
A heterogeneous graph $G = (V, E)$ is associated with a node type mapping function $\Phi: V \rightarrow \mathcal{A}$ and an edge type mapping function $\Psi: E \rightarrow \mathcal{R}$ and $|\mathcal{A}| > 1$ and/or $|\mathcal{R}| > 1$. 
\end{definition}

\vspace{-0.2in}
\begin{definition}[Dynamic graphs]
A graph $G = (V, E)$ is a dynamic graph where $V = (v, t_s, t_e)$ with $t_s$, $t_e$ are respectively the start and
end timestamps for the vertex existence (with $t_s \leq t_e$); $E = {(u, v, t_s, t_e)}$ with $u, v \in V$ and $t_s$, $t_e$ are respectively the start and end timestamps for the edge existence (with $t_s \leq t_e$).
\end{definition}

\vspace{-0.2in}
\begin{problem}[\textbf{Graph embedding}]
Given a graph $G = (V, E)$, and a predefined embedding dimensionality $d$ where $d \ll |V|$. Graph embedding aims to convert $G$ into a $d$-dimensional space $\mathbb{R}^d$, where the information and proprieties of $G$ are well preserved as much as possible. 
\end{problem}
\vspace{-0.1in}

In the following section, we provide the taxonomy of graph embedding methods based on the graph settings and embedding techniques, respectively.

\section{Taxonomy of Graph Embedding Models}
\vspace{-0.1in}

\begin{figure}[]
\centering
     \includegraphics[width=1\columnwidth]{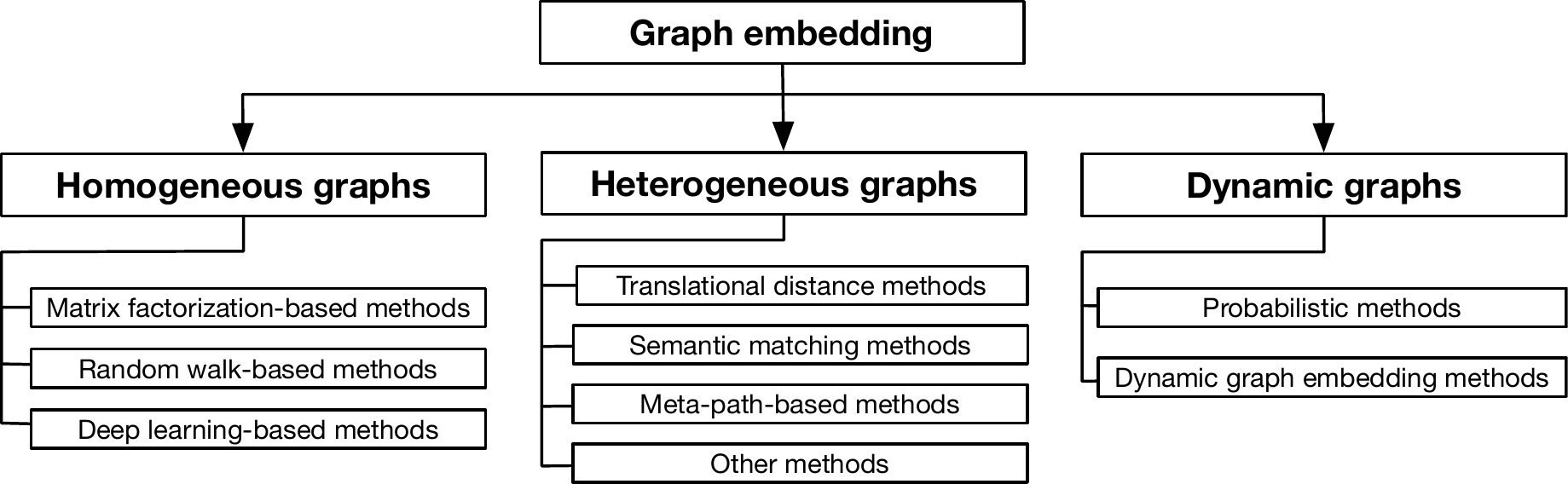}
    \caption{Taxonomy of graph embedding models.}
    \label{fig:taxonomy}
\end{figure}

As shown in Figure~\ref{fig:taxonomy}, in this section, according to the graph settings, we introduce \textit{homogeneous graph embedding models}, \textit{heterogeneous graph embedding models} and \textit{dynamic graph embedding models} as follows.

\vspace{-0.1in}
\subsection{Homogeneous Graph Embedding Models}
\vspace{-0.1in}

In the literature, there are three main types of homogeneous graph embedding methods, i.e., \textit{matrix factorization-based methods}, \textit{random walk-based methods} and \textit{deep learning-based methods}.

\noindent\textbf{Matrix factorization-based methods.} Matrix factorization-based methods, inspired by classic techniques for dimensionality reduction, use the form of a matrix to represent the graph properties, e.g., node pairwise similarity.
Generally, there are two types of matrix factorization to compute the node embedding, i.e., \textit{node proximity matrix} and \textit{graph Laplacian eigenmaps}. 

For node proximity matrix factorization methods, they usually approximate node proximity into a low dimension and the objective of preserving node proximity is to minimize the approximation loss $||W-UX^{T}||$, where $W$ is the node proximity matrix, $X$ is the embedding for context nodes and embedding $U$ can be computed using this loss function. 
Actually, there are many other solutions to approximate this loss function, such as low rank matrix factorization, regularized Gaussian matrix factorization, etc. 
For graph Laplacian eigenmaps factorization methods, the assumption is that the graph property can be interpreted as the similarity of pairwise nodes. Thus, to obtain a good representation, the normal operation is that a larger penalty will be given if two nodes with higher similarity are far embedded. The optimal embedding $U^*$ can be computed by using the objective function~(\ref{eq:reducedL}):
\begin{equation}
\label{eq:reducedL}
U^* =  \mathop{\arg\min}_{U^TDU = 1}U^TLU = \arg \min \frac{U^TLU}{U^TDU} = \arg \max \frac{U^TWU}{U^TDU},
\end{equation}
where $L = D - W$ is the graph Laplacian. $D$ is the diagonal matrix and $D_{ii} = \sum_{j} W_{ji}$. 
There are many works using graph Laplacian-based methods and they mainly differ from how they calculate the pairwise node similarity $W_{ij}$. 
For example, BANE~\cite{yang2018binarized} defines a new Weisfeiler-Lehman proximity matrix
to capture data dependence between edges and attributes; then based on this matrix, BANE learns the node embeddings by formulating a new Weisfiler-Lehman matrix factorization. Recently, NetMF~\cite{qiu2018network} unifies state-of-the-art approaches into a matrix factorization framework with close forms. 

\noindent\textbf{Random walk-based methods.}
Random walk-based methods have been widely used to approximate many properties in the graph including node centrality and similarity. They are more useful when the graph can only partially be observed, or the graph is too large to measure. Two widely recognized random walk-based methods have been proposed, i.e., DeepWalk~\cite{perozzi2014deepwalk} and node2vec~\cite{grover2016node2vec}. Concretely, DeepWalk considers the paths as sentences and implements an NLP model to learn node embeddings. Compared to DeepWalk, node2vec introduces a trade-off strategy using breadth-first and depth-first search to perform a biased random walk. In recent year, there are still many random walk-based papers working on improving performance. For example, AWE~\cite{ivanov2018anonymous} uses a recently developed method called \textit{anonymous walks}, i.e., an anonymized version of the random walk-based method providing characteristic graph traits and are capable to exactly reconstruct network proximity of a node. AttentionWalk~\cite{abu2018watch} uses the softmax to learn a free-form context distribution in a random walk; then the learned attention parameters guide the random walk, by allowing it to focus more on short or long term dependencies when optimizing an upstream objective. BiNE~\cite{gao2018bine} proposes methods for bipartite graph embedding by performing biased random walks. Then they generate vertex sequences that can well preserve the long-tail distribution of vertices in original bipartite graphs.

\noindent\textbf{Deep learning-based methods.}
Deep learning has shown outstanding performance in a wide variety of research fields. SDNE~\cite{wang2016structural} applies a deep autoencoder to model non-linearity in the graph structure. DNGR~\cite{cao2016deep} learns deep low-dimensional vertex representations, by using the stacked denoising autoencoders on the high-dimensional matrix representations. Furthermore, Graph Convolutional Network (GCN)~\cite{kipf2016semi} introduces a well-behaved layer-wise propagation rule for the neural network model which operates directly on graphs in Equation~(\ref{eq:gcn}):
\begin{equation}
\label{eq:gcn}
H^{l+1} = \sigma(\hat{D}^{-\frac{1}{2}}\hat{A}\hat{D}^{-\frac{1}{2}}H^{l}W^{l}),
\end{equation}
with $\hat{A} = A + I$, where $A$ and $I$ is the adjacency and identity matrix and $\hat{D}$ is the diagonal degree matrix of $\hat{A}$. $w^l$ is a weight matrix for the $l$-th neural network layer and $\sigma(\cdot)$ is a non-linear activation function like the \textbf{ReLU}. $H^{l}$  and $H^{l+1}$ are the input and output for layer $l$ and layer $l+1$, respectively.
Another important work is Graph Attention Network (GAT)~\cite{velivckovic2017graph}, which leverages masked self-attentional layers to address the shortcomings of prior graph convolution-based methods. Specifically, as shown in Equation~(\ref{eq:attention}):
\begin{equation}
\label{eq:attention}
     \alpha_{ij} = \frac{\exp(e_{ij})}{\sum_{k\in\mathcal{N}_i}\exp(e_{ik})}, {\rm \ \ where \ \ } e_{ij} = a(\vec{h}_i, \vec{h}_j).
\end{equation}
$\mathcal{N}_i$ is the neighbors of node $i$. GAT computes norm$i$alized coefficients $\alpha_{ij}$ using the softmax function across different neighborhoods by a byproduct of an attentional mechanism across node pairs. To stabilize the learning process of self-attention, GAT uses multi-head attention to replicate $K$ times of learning phases, and outputs are feature-wise aggregated (typically by concatenating or adding), as shown in Equation~(\ref{eq:gat}):
\begin{equation}
\label{eq:gat}
h_i = ||^K_{k = 1}\sigma(\sum_{j \in \mathcal{N}_j}\alpha_{ij}W^kh_j),
\end{equation}
where $\alpha^k_{ij}$ and $W_k$ are the attention coefficients and the weight matrix specifying the linear transformation of the $k$-th replica.
Recently, HGCN~\cite{chami2019hyperbolic} and ATTH~\cite{chami2020low} use hyperbolic model to embed hierarchical graph structure with less distortion.

\vspace{-0.1in}
\subsection{Heterogeneous Graph Embedding Models}
\vspace{-0.1in}

The heterogeneity in both graph structures and node attributes makes it challenging for the graph embedding task to encode their diverse and rich information. In this section, we will introduce \textit{translational distance methods} and \textit{semantic matching methods}, which try to address the above issue by constructing different energy functions. Furthermore, we will introduce \textit{meta-path-based methods} that use different strategies to capture graph heterogeneity.

\noindent\textbf{Translational distance methods.}
The first work of translation distance models is TransE~\cite{bordes2013translating}. The basic idea of the translational distance models is, for each observed fact $(h,r,t)$ representing head entity $h$ having a relation $r$ with tail entity $t$, to learn a good graph representation such that $h$ and $t$ are closely connected by relation $r$ in low dimensional embedding space, i.e., \textbf{h} + \textbf{r} $\approx$ \textbf{t} using geometric notations. Here \textbf{h}, \textbf{r} and \textbf{t} are embedding vectors for entities $h$, $t$ and relation $r$, respectively.
The energy function of TransE is defined as $f_r(h, t) = ||\textbf{h} + \textbf{r} - \textbf{t}||^2_2$. The margin-based objective funtion of TransE is shown in Equation~(\ref{eq:objF}):
\begin{equation}
\label{eq:objF}
\mathcal{L} = \sum_{(h,r,t) \in S} \sum_{(h',r,t') \in S'} \max(0, f_r(h, t) - f_r(h', t') + margin),
\end{equation}
where $S$ denotes the set containing the true facts, e.g., $(h,r,t)$, and $S'$ is the set of false triplets, e.g., $(h',r,t')$, that are not observed in the knowledge graphs. Please note that the energy function $f_r$ here can be viewed as the distance score of the embedding of entities $h$ and $t$ in terms of relation $r$. To further improve TransE model and address its inadequacies, many recent works have been developed. For example, RotatE~\cite{sun2019rotate} defines each relation as a rotation from the source entity to the target entity in the complex vector space. QuatE~\cite{zhang2019quaternion} computes node embedding vectors in the hypercomplex space with three imaginary components, as opposed to the standard complex space with a single real component and imaginary component. 
MuRP~\cite{balazevic2019multi} is a hyperbolic embedding method that embeds multi-relational data in the Poincar{\'e} ball model of hyperbolic space, which can well perform in hierarchical and scale-free graphs.

\noindent\textbf{Semantic matching methods.}
Semantic matching models exploit similarity-based scoring functions. They measure plausibility of facts by matching latent semantics of entities and relations embodied in their representations. Targetting the observed fact $(h,r,t)$, RESCAL~\cite{nickel2011three} embeds each entity with a vector to capture its latent semantics and each relation with a matrix to model pairwise interactions between latent factors. Equation~(\ref{eq:RESCAL}) defines the energy function:
\begin{equation}
\label{eq:RESCAL}
f_r(h, t) = \textbf{h}^T M_r \textbf{t} = \sum_{i = 0}^{d-1} \sum_{j = 0}^{d - 1} [M_{r}]_{ij} \cdot \textbf{h}_i \cdot \textbf{t}_j,
\end{equation}
where $M_r$ is a matrix associated with the relation. HolE~\cite{nickel2016holographic} deals with directed graphs and composes head entity and tail entity by their circular correlation, which achieves a better performance than RESCAL. There are other works trying to extend or simplify RESCAL, e.g., DistMult~\cite{yang2014embedding}, ComplEx~\cite{trouillon2016complex}, ANALOGY~\cite{liu2017analogical}. Other direction of semantic matching methods is to fuse neural network architecture by considering embedding as the input layer and energy function as the output layer. For instance, SME model~\cite{bordes2014semantic} first imputs embeddings of entities and relations in the input layer. The relation $r$ is then combined with the head entity $h$ to get $g_{left}(h, r) = M_1\textbf{h} + M_2\textbf{r} + \textbf{b}_h$, and with the tail entity $t$ to get $g_{right}(t, r) = M_3\textbf{t} + M_4\textbf{r}+\textbf{b}_t$ in the hidden layer. The score function is defined as $f_r(h,t) = g_{left}(h,r)^T \cdot g_{right}(t,r)$. There are other semantic matching methods using neural network architecture, e.g., NTN~\cite{socher2013reasoning}, MLP~\cite{dong2014knowledge}.

\noindent\textbf{Meta-path-based methods.}
Generally, a meta-path is an ordered path that consists of node types and connects via edge types defined on the graph schema, e.g., $A_1 \stackrel{R_1}{\longrightarrow} A_2 \cdots \stackrel{R_{l-1}}{\longrightarrow} A_{l}$, which describes a composite relation between node types $A_1$, $A_2$, $\cdots$, $A_l$ and edge types $R_1$, $\cdots$, $R_{l-1}$.  
Thus, meta-paths can be viewed as high-order proximity between two nodes with specific semantics. A set of recent works have been proposed. Metapath2vec~\cite{dong2017metapath2vec} computes node embeddings by feeding metapath-guided random walks to a skip-gram\cite{mikolov2013efficient} model. HAN~\cite{wang2019heterogeneous} learns meta-path-oriented node embeddings from different meta-path-based graphs converted from the original heterogeneous graph and leverages the attention mechanism to combine them into one vector representation for each node. HERec~\cite{shi2018heterogeneous} learns node embeddings by applying DeepWalk~\cite{perozzi2014deepwalk} to the meta-path-based homogeneous graphs for recommendation. MAGNN~\cite{fu2020magnn} comprehensively considers three main components to achieve the state-of-the-art performance. Concretely, MAGNN~\cite{fu2020magnn} fuses the node content transformation to encapsulate node attributes, the intra-metapath aggregation to incorporate intermediate semantic nodes, and the inter-metapath aggregation to combine messages from multiple metapaths. 

\noindent\textbf{Other methods.}
LANE~\cite{huang2017label} constructs proximity matrices by incorporating label information, graph topology, and learns embeddings while preserving their correlations based on Laplacian matrix. EOE~\cite{xu2017embedding} aims to embed the graph coupled by two non-attribute graphs. 
In EOE, latent features encode not only intra-network edges, but also inter-network ones. To tackle the challenge of heterogeneity of two graphs, the EOE incorporates a harmonious embedding matrix to further embed the embeddings. Inspired by generative adversarial network models,  HeGAN~\cite{hu2019adversarial} is designed to be relation-aware in order to capture the rich semantics on heterogeneous graphs and further trains a discriminator and a generator in a minimax game to generate robust graph embeddings. 

\vspace{-0.1in}
\subsection{Dynamic Graph Embedding Models}
\vspace{-0.1in}

In practice, graphs are always evolving over time. Recently, much attention is paid to graph embedding for dynamic graphs. In this section, we will briefly introduce some typical general models as follows.

\noindent\textbf{Probabilistic models.}
In generative probabilistic models, \textit{Dynamic latent space models} and \textit{Dynamic stochastic block models} are two main types within. Latent space models model every node with an unobserved feature vector. An edge between two nodes is then formed conditionally independent of all other pairs of nodes. The latent features are changed over time. Such models are flexible and require fitting of parameters with Markov chain Monte Carlo methods that scale up to only a few hundred nodes~\cite{junuthula2016evaluating}. Stochastic block models divide nodes into blocks (classes), where nodes within a block are assumed to have identical statistical properties. An edge between two nodes is formed independently of all other pairs of nodes with a probability dependents only on the blocks of the two nodes, giving the adjacency matrix of blocks corresponding to pairs of blocks.

\noindent\textbf{Dynamic graph embedding methods.}
In dynamic graph embedding methods, there are mainly three types of methods, i.e., \textit{tensor decomposition-based methods}, \textit{random walk-based methods}, \textit{deep learning-based methods}, which are actually inspired from those for homogeneous graphs. Tensor decomposition is analogous to matrix factorization where the additional dimension is time. As for random walk-based methods for dynamic graphs, they are generally extensions of random walk-based embedding methods for static graphs or they apply temporal random walks. Furthermore, deep learning models for dynamic graphs mainly contain two types of models: temporal restricted Boltzmann machines and dynamic graph neural networks. For detailed analysis, please refer to the survey over dynamic graph embedding in~\cite{skarding2020foundations,kazemi2020representation}.

\section{Applications and Tasks in Biomedical Domain}
\vspace{-0.1in}

\subsection{Biomedical datasets}
\vspace{-0.1in}

We first summarize some commonly used biomedical  datasets in Table~\ref{tb:data}, where the columns are: average number of nodes/edges, dimensionality of node features, number of node classes, and graphs, respectively.

\begin{table}[htbp]      
\centering
\caption{Datasets Statistics}
\label{tb:data}
\vspace{-0.1in}
    \begin{tabular}{c|c|c|c|c|c|c}    
        \hline
        {\textbf{Dataset}} & {\textbf{avg. $|V|$}} & {\textbf{avg. $|E|$}} & {\textbf{Features}} & {\textbf{Classes}} & {\textbf{Graphs}} & {\textbf{Graph Type}}\\        
        \hline
        \hline
        {PubMed-diabetes} & {19,717.00} & {44,338.00} & {500} & {3} & {1} & {Citation Graph} \\  
        \hline
        {PPI} & {2,372.67} & {34,113.17} & {50} & {121} & {24} & {Bio-chemical Graph} \\ 
        \hline
        {MUTAG} & {17.93} & {19.79} & {7} & {2} & {188} & {Bio-chemical Graph} \\ 
        \hline
        {NCI-1} & {29.87} & {32.30} & {37} & {2} & {4,110} & {Bio-chemical Graph} \\ 
        \hline
        {NCI-33} & {30.20} & {-} & {29} & {-} & {2,843} & {Bio-chemical Graph} \\ 
        \hline
        {NCI-83} & {29.50} & {-} & {28} & {-} & {3,867} & {Bio-chemical Graph} \\ 
        \hline
        {NCI-109} & {29.60} & {-} & {38} & {-} & {4,127} & {Bio-chemical Graph} \\ 
        \hline
        {DD} & {284.31} & {715.65} & {82} & {2} & {1,178} & {Bio-chemical Graph} \\ 
        \hline
        {PROTEINS} & {39.06} & {72.81} & {4} & {2} & {1,113} & {Bio-chemical Graph} \\ 
        \hline
        {ENZYMES} & {32.46} & {63.14} & {6} & {6} & {600} & {Biological Graph} \\ 
        \hline
        \end{tabular}
\end{table}

PubMed-diabetes\footnote{https://linqs.soe.ucsc.edu/data} is a citation graph consists of scientific publications and citations pertaining to diabetes. PPI\footnote{http://snap.stanford.edu/graphsage/ppi.zip} contains 24 graphs including protein-protein interactions of different organisms such as Homo sapiens, Mus musculus, etc. MUTAG\footnote{https://ls11-www.cs.uni-dortmund.de/people/morris/graphkerneldatasets\label{ds:bio}} dataset contains nitro compounds which are divided into two classes according to their mutagenic effect on a bacterium. 
NCI-\{1, 33, 83, 109\}\cite{pan2013graph} contains chemical compounds which are screened for activity against non-small cell cancer of lung, melanoma, breast and ovarian, respectively.
DD\footnote{https://chrsmrrs.github.io/datasets/docs/datasets/\label{ds:dd}} and PROTEINS\textsuperscript{\ref{ds:dd}} are two datasets that represent proteins as graphs which labels are enzymes and non-enzymes and  
ENZYMES\cite{xinyi2018capsule} is a biological dataset.

\vspace{-0.1in}
\subsection{Applications and Tasks}
\vspace{-0.1in}

In recent years, graph embedding methods have been applied in biomedical data science. In this section, we will introduce  some main biomedical applications of applying graph embedding techniques, including \textit{pharmaceutical data analysis}, \textit{multi-omics data analysis} and \textit{clinical data analysis}.

\noindent\textbf{Pharmaceutical data analysis.}
Generally, there are two main types of applications for  pharmaceutical data analysis, i.e., (i)~\textit{drug repositioning} and (ii) \textit{adverse drug reaction analysis}.

(i) Drug repositioning usually aims to predict unknown drug-target or drug-disease interactions. 
Recently, DTINet~\cite{luo2017network} generates drug and target-protein embedding by separately performing random walk with restart on heterogeneous biomedical graphs. Then DTINet projects drugs into the embedding space of target proteins and made predictions based on geometric proximity. Other studies over drug repositioning focused on predicting drug disease associations. For instance,  Dai et al.~\cite{dai2015matrix}  first embed genes by applying eigenvalue decomposition to a gene-gene interaction graph and calculated genomic representations for drugs and diseases from the gene embedding vectors.  Wang et al.~\cite{wang2017large}  proposed to detect unknown drug-disease interactions from the medical literature by fusing NLP and graph embedding techniques. 
(ii) An adverse drug reaction (ADR) is defined as any undesirable drug effect out of its desired therapeutic effects that occur at a usual dosage, which now is the center of drug development before a drug is launched on the clinical trial. 

\noindent\textbf{Multi-omics data analysis.}
The main aim of multi-omics is to study structures, functions, and dynamics of organism molecules. Fortunately, graph embedding now becomes a valuable tool to analyze relational data in omics. Concretely, the computation tasks included in multi-omics data analysis are mainly about (i)~\textit{genomics}, (ii)~\textit{proteomics} and (iii)~\textit{transcriptomics}.

(i) Works of graph embedding used in genomics data analysis usually try to decipher biology from genome sequences and related data. For example, based on gene-gene interaction data, a recent work~\cite{li2017network} extends the graph embedding method, i.e., LINE, over two bipartite graphs, Cell-ContexGene and Gene-ContexGene networks, and then proposes SCRL to address representation learning for single cell RNA-seq data, which outperforms traditional dimensional reduction methods according to the experimental results. 
(ii) As we have introduced before, PPIs play key roles in most cell functions. Graph embedding has also been introduced to PPI graphs for proteomics data analysis, such as assessing and predicting PPIs or predicting protein functions, etc. 
Recently, ProSNet~\cite{wang2017prosnet} has been proposed for protein function prediction. 
In this model, they introducing DCA to a heterogeneous molecular graph and further use the meta-path-based methods to modify DCA for preserving heterogeneous structural information. Thanks to the proposed embedding methods for such heterogeneous graphs, their experimental prediction performance was greatly improved.
(iii) As for transcriptomics study, the focus is to analyze an organism's transcriptome. For instance, Identifying miRNA-disease associations now becomes an important topic of pathogenicity; while graph embedding now provides a useful tool to involve in transcriptomics for prediction of miRNA-disease associations. To predict new associations, CMFMDA~\cite{shen2017mirna} introduces matrix factorization methods to the bipartite miRNA-disease graph for graph embedding. Besides, Li et al.~\cite{li2017predicting} proposed a method by using DeepWalk to embed the bipartite miRNA-disease network. Their experimental results demonstrate that, by preserving both local and global graph topology, DeepWalk can result in significant improvements in association prediction for miRNA-disease graphs.

\noindent\textbf{Clinical data analysis.}
Graph embedding techniques have been applied to clinic data, such as electronic medical records (EMRs), electronic health records (EHRs) and medical knowledge graph, providing useful assistance and support for clinicians in recent clinic development.

EMRs and EHRs are heterogeneous graphs that comprehensively include medical and clinical information from patients, which provide opportunities for graph embedding techniques to make medical research and clinical decision. To address the heterogeneity of EMRs and EHRs data, GRAM~\cite{choi2017gram} learns EHR representation with the help of
hierarchical information inherent to medical ontologies. ProSNet~\cite{huang2018visage} constructs a biomedical knowledge graph to learn the embeddings of medical entities. The proposed method is used to visualize the Parkinson’s disease data set.
Conducting medical knowledge graph is of great importance and attention recently. For instance, analogous to TransE, Zhao et al.~\cite{zhao2017contextcare} defined energy function by considering the relation between the symptoms of patients and diseases as a translation vector to further learn the representation of medical forum data. Then a new method is proposed to learn embeddings of medical entities in the medical knowledge graph, based on the energy functions of RESCAL and TransE~\cite{zhao2018emr}. In addition, Wang et al.~\cite{wang2017safe} constructed objective function by using both the energy function of TransR and LINE’s 2nd-order proximity measurement to learn embeddings from a heterogeneous medical knowledge graph to further recommend proper medicine to patients.

\section{Conclusion}
\vspace{-0.1in}

Graph embedding methods aim to learn compact and informative representations for graph analysis and thus provide a powerful opportunity to solve the traditional graph-based machine learning problems both effectively and efficiently. With the rapid development of relational data in the biomedical data domain, applying graph embedding techniques now draws much attention in numerous biomedical applications. 
However, as we have reviewed in this survey, the capability of graph embedding for biomedical graph analysis has not been fully explored. There may exist many issues associated with the biomedical data that may bring challenges to biomedical graph embedding tasks. For example, biomedical data quality could be not well structured; knowledge and information from biomedical domain or health care records could be complicated, compared to the general domain. In this survey, we introduce recent developments and trends of different graph embedding methods. By carefully summarizing biomedical applications with graph embedding methods, we provide more perspectives over this emerging research domain for better improvement in human health care.

%
%
\bibliographystyle{splncs04}
\vspace{-0.05in}
\bibliography{ref}

\begin{thebibliography}{10}
\providecommand{\url}[1]{\texttt{#1}}
\providecommand{\urlprefix}{URL }
\providecommand{\doi}[1]{https://doi.org/#1}

\bibitem{abu2018watch}
Abu-El-Haija, S., Perozzi, B., Al-Rfou, R., Alemi, A.A.: Watch your step:
  Learning node embeddings via graph attention. In: NeurIPS. pp. 9180--9190
  (2018)

\bibitem{albert2005scale}
Albert, R.: Scale-free networks in cell biology. Journal of cell science
  (2005)

\bibitem{balazevic2019multi}
Balazevic, I., Allen, C., Hospedales, T.: Multi-relational poincar{\'e} graph
  embeddings. In: NeurIPS. pp. 4465--4475 (2019)

\bibitem{barabasi2011network}
Barab{\'a}si, A.L., Gulbahce, N., Loscalzo, J.: Network medicine: a
  network-based approach to human disease. Nature reviews genetics
  \textbf{12}(1),  56--68 (2011)

\bibitem{bordes2014semantic}
Bordes, A., Glorot, X., Weston, J., Bengio, Y.: A semantic matching energy
  function for learning with multi-relational data. ML  \textbf{94}(2),
  233--259 (2014)

\bibitem{bordes2013translating}
Bordes, A., Usunier, N., Garcia-Duran, A., Weston, J., Yakhnenko, O.:
  Translating embeddings for modeling multi-relational data. In: NeurIPS. pp.
  2787--2795 (2013)

\bibitem{cai2018comprehensive}
Cai, H., Zheng, V.W., Chang, K.C.C.: A comprehensive survey of graph embedding:
  Problems, techniques, and applications. TKDE  \textbf{30}(9),  1616--1637
  (2018)

\bibitem{cao2016deep}
Cao, S., Lu, W., Xu, Q.: Deep neural networks for learning graph
  representations. In: AAAI (2016)

\bibitem{chami2020machine}
Chami, I., Abu-El-Haija, S., Perozzi, B., R{\'e}, C., Murphy, K.: Machine
  learning on graphs: A model and comprehensive taxonomy. arXiv preprint
  arXiv:2005.03675  (2020)

\bibitem{chami2020low}
Chami, I., Wolf, A., Juan, D.C., Sala, F., Ravi, S., R{\'e}, C.:
  Low-dimensional hyperbolic knowledge graph embeddings. ACL  (2020)

\bibitem{chami2019hyperbolic}
Chami, I., Ying, Z., R{\'e}, C., Leskovec, J.: Hyperbolic graph convolutional
  neural networks. In: NeurIPS. pp. 4868--4879 (2019)

\bibitem{chen2020efficient}
Chen, Y., Zhang, J., Fang, Y., Cao, X., King, I.: Efficient community search
  over large directed graph: An augmented index-based approach. In: IJCAI. pp.
  3544--3550 (2020)

\bibitem{choi2017gram}
Choi, E., Bahadori, M.T., Song, L., Stewart, W.F., Sun, J.: Gram: graph-based
  attention model for healthcare representation learning. In: SIGKDD (2017)

\bibitem{dai2015matrix}
Dai, W., Liu, X., Gao, Y., Chen, L., Song, J., Chen, D., Gao, K., Jiang, Y.,
  Yang, Y., Chen, J., et~al.: Matrix factorization-based prediction of novel
  drug indications by integrating genomic space. CMMM  \textbf{2015} (2015)

\bibitem{dong2014knowledge}
Dong, X., Gabrilovich, E., Heitz, G., Horn, W., Lao, N., Murphy, K., Strohmann,
  T., Sun, S., Zhang, W.: Knowledge vault: A web-scale approach to
  probabilistic knowledge fusion. In: SIGKDD. pp. 601--610 (2014)

\bibitem{dong2017metapath2vec}
Dong, Y., Chawla, N.V., Swami, A.: metapath2vec: Scalable representation
  learning for heterogeneous networks. In: SIGKDD. pp. 135--144 (2017)

\bibitem{fu2020magnn}
Fu, X., Zhang, J., Meng, Z., King, I.: Magnn: Metapath aggregated graph neural
  network for heterogeneous graph embedding. In: WWW. pp. 2331--2341 (2020)

\bibitem{gao2018bine}
Gao, M., Chen, L., He, X., Zhou, A.: Bine: Bipartite network embedding. In:
  SIGIR. pp. 715--724 (2018)

\bibitem{goyal2018graph}
Goyal, P., Ferrara, E.: Graph embedding techniques, applications, and
  performance: A survey. Knowledge-Based Systems  \textbf{151},  78--94 (2018)

\bibitem{grover2016node2vec}
Grover, A., Leskovec, J.: node2vec: Scalable feature learning for networks. In:
  SIGKDD. pp. 855--864 (2016)

\bibitem{hu2019adversarial}
Hu, B., Fang, Y., Shi, C.: Adversarial learning on heterogeneous information
  networks. In: SIGKDD. pp. 120--129 (2019)

\bibitem{huang2018visage}
Huang, E.W., Wang, S., Zhai, C.: Visage: Integrating external knowledge into
  electronic medical record visualization. In: PSB. pp. 578--589. World
  Scientific (2018)

\bibitem{huang2017label}
Huang, X., Li, J., Hu, X.: Label informed attributed network embedding. In:
  WSDM. pp. 731--739 (2017)

\bibitem{ivanov2018anonymous}
Ivanov, S., Burnaev, E.: Anonymous walk embeddings. arXiv:1805.11921  (2018)

\bibitem{junuthula2016evaluating}
Junuthula, R.R., Xu, K.S., Devabhaktuni, V.K.: Evaluating link prediction
  accuracy in dynamic networks with added and removed edges. In:
  BDCloud-SocialCom-SustainCom. pp. 377--384. IEEE (2016)

\bibitem{kazemi2020representation}
Kazemi, S.M., Goel, R., Jain, K., Kobyzev, I., Sethi, A., Forsyth, P., Poupart,
  P.: Representation learning for dynamic graphs: A survey. Journal of Machine
  Learning Research  \textbf{21}(70),  1--73 (2020)

\bibitem{kipf2016semi}
Kipf, T.N., Welling, M.: Semi-supervised classification with graph
  convolutional networks. arXiv:1609.02907  (2016)

\bibitem{li2017predicting}
Li, G., Luo, J., Xiao, Q., Liang, C., Ding, P., Cao, B.: Predicting
  microrna-disease associations using network topological similarity based on
  deepwalk. IEEE Access  \textbf{5},  24032--24039 (2017)

\bibitem{li2017network}
Li, X., Chen, W., Chen, Y., Zhang, X., Gu, J., Zhang, M.Q.: Network
  embedding-based representation learning for single cell rna-seq data. Nucleic
  acids research  \textbf{45}(19),  e166--e166 (2017)

\bibitem{liu2017analogical}
Liu, H., Wu, Y., Yang, Y.: Analogical inference for multi-relational
  embeddings. In: ICML. pp. 2168--2178 (2017)

\bibitem{luo2017network}
Luo, Y., Zhao, X., Zhou, J., Yang, J., Zhang, Y., Kuang, W., Peng, J., Chen,
  L., Zeng, J.: A network integration approach for drug-target interaction
  prediction and computational drug repositioning from heterogeneous
  information. Nature communications  \textbf{8}(1),  1--13 (2017)

\bibitem{mikolov2013efficient}
Mikolov, T., Chen, K., Corrado, G., Dean, J.: Efficient estimation of word
  representations in vector space. In: {ICLR} (Workshop Poster) (2013)

\bibitem{nickel2016holographic}
Nickel, M., Rosasco, L., Poggio, T.: Holographic embeddings of knowledge
  graphs. In: AAAI (2016)

\bibitem{nickel2011three}
Nickel, M., Tresp, V., Kriegel, H.P.: A three-way model for collective learning
  on multi-relational data. In: ICML. vol.~11, pp. 809--816 (2011)

\bibitem{pan2013graph}
Pan, S., Zhu, X., Zhang, C., Philip, S.Y.: Graph stream classification using
  labeled and unlabeled graphs. In: ICDE. pp. 398--409. IEEE (2013)

\bibitem{perozzi2014deepwalk}
Perozzi, B., Al-Rfou, R., Skiena, S.: Deepwalk: Online learning of social
  representations. In: SIGKDD. pp. 701--710 (2014)

\bibitem{qiu2018network}
Qiu, J., Dong, Y., Ma, H., Li, J., Wang, K., Tang, J.: Network embedding as
  matrix factorization: Unifying deepwalk, line, pte, and node2vec. In: WSDM
  (2018)

\bibitem{shen2017mirna}
Shen, Z., Zhang, Y.H., Han, K., Nandi, A.K., Honig, B., Huang, D.S.:
  mirna-disease association prediction with collaborative matrix factorization.
  Complexity  (2017)

\bibitem{shi2018heterogeneous}
Shi, C., Hu, B., Zhao, W.X., Philip, S.Y.: Heterogeneous information network
  embedding for recommendation. TKDE  \textbf{31}(2),  357--370 (2018)

\bibitem{shickel2017deep}
Shickel, B., Tighe, P.J., Bihorac, A., Rashidi, P.: Deep ehr: a survey of
  recent advances in deep learning techniques for electronic health record
  (ehr) analysis. IEEE journal of biomedical and health informatics
  \textbf{22}(5),  1589--1604 (2017)

\bibitem{skarding2020foundations}
Skarding, J., Gabrys, B., Musial, K.: Foundations and modelling of dynamic
  networks using dynamic graph neural networks: A survey. arXiv:2005.07496
  (2020)

\bibitem{socher2013reasoning}
Socher, R., Chen, D., Manning, C.D., Ng, A.: Reasoning with neural tensor
  networks for knowledge base completion. In: NeurIPS. pp. 926--934 (2013)

\bibitem{su2020network}
Su, C., Tong, J., Zhu, Y., Cui, P., Wang, F.: Network embedding in biomedical
  data science. Briefings in bioinformatics  \textbf{21}(1),  182--197 (2020)

\bibitem{sun2019rotate}
Sun, Z., Deng, Z., Nie, J., Tang, J.: Rotate: Knowledge graph embedding by
  relational rotation in complex space. In: {ICLR} (Poster). OpenReview.net
  (2019)

\bibitem{trouillon2016complex}
Trouillon, T., Welbl, J., Riedel, S., Gaussier, {\'E}., Bouchard, G.: Complex
  embeddings for simple link prediction. ICML (2016)

\bibitem{velivckovic2017graph}
Veli{\v{c}}kovi{\'c}, P., Cucurull, G., Casanova, A., Romero, A., Lio, P.,
  Bengio, Y.: Graph attention networks. arXiv:1710.10903  (2017)

\bibitem{wang2016structural}
Wang, D., Cui, P., Zhu, W.: Structural deep network embedding. In: SIGKDD. pp.
  1225--1234 (2016)

\bibitem{wang2017safe}
Wang, M., Liu, M., Liu, J., Wang, S., Long, G., Qian, B.: Safe medicine
  recommendation via medical knowledge graph embedding. arXiv:1710.05980
  (2017)

\bibitem{wang2017large}
Wang, P., Hao, T., Yan, J., Jin, L.: Large-scale extraction of drug--disease
  pairs from the medical literature. Journal of the AIST  \textbf{68}(11),
  2649--2661 (2017)

\bibitem{wang2017prosnet}
Wang, S., Qu, M., Peng, J.: Prosnet: Integrating homology with molecular
  networks for protein function prediction. In: PSB. pp. 27--38. World
  Scientific (2017)

\bibitem{wang2019heterogeneous}
Wang, X., Ji, H., Shi, C., Wang, B., Ye, Y., Cui, P., Yu, P.S.: Heterogeneous
  graph attention network. In: WWW. pp. 2022--2032 (2019)

\bibitem{xinyi2018capsule}
Xinyi, Z., Chen, L.: Capsule graph neural network. In: {ICLR} (Poster).
  OpenReview.net (2019)

\bibitem{xu2017embedding}
Xu, L., Wei, X., Cao, J., Yu, P.S.: Embedding of embedding (eoe) joint
  embedding for coupled heterogeneous networks. In: WSDM. pp. 741--749 (2017)

\bibitem{yang2014embedding}
Yang, B., Yih, W.t., He, X., Gao, J., Deng, L.: Embedding entities and
  relations for learning and inference in knowledge bases. arXiv:1412.6575
  (2014)

\bibitem{yang2018binarized}
Yang, H., Pan, S., Zhang, P., Chen, L., Lian, D., Zhang, C.: Binarized
  attributed network embedding. In: ICDM. pp. 1476--1481. IEEE (2018)

\bibitem{zhang2019quaternion}
Zhang, S., Tay, Y., Yao, L., Liu, Q.: Quaternion knowledge graph embeddings.
  In: NeurIPS. pp. 2731--2741 (2019)

\bibitem{zhao2018emr}
Zhao, C., Jiang, J., Guan, Y., Guo, X., He, B.: {EMR}-based medical knowledge
  representation and inference via markov random fields and distributed
  representation learning. Artificial intelligence in medicine  \textbf{87},
  49--59 (2018)

\bibitem{zhao2017contextcare}
Zhao, S., Jiang, M., Yuan, Q., Qin, B., Liu, T., Zhai, C.: Contextcare:
  Incorporating contextual information networks to representation learning on
  medical forum data. In: IJCAI. pp. 3497--3503 (2017)

\end{thebibliography}

\end{document}